\pdfoutput=1
\documentclass[journal]{IEEEtran}

\usepackage{amsmath}
\usepackage{algorithm}
\usepackage{algpseudocode}
\usepackage[pdftex]{graphicx}
\usepackage{tabularx}
\usepackage{booktabs}
\usepackage{makecell}
\usepackage{float}
\usepackage{soul}
\usepackage[numbers,sort&compress]{natbib}
\usepackage{hyperref}
\usepackage{xcolor}
\usepackage{subcaption}
\usepackage{ragged2e}
\usepackage{amsfonts}
\usepackage{amssymb}

\title{\vspace{-10mm}\rule{\linewidth}{0.4mm}\\[2mm]
      Domain Specific Data Distillation and Multi-modal Embedding Generation
       \\[2mm]\rule{\linewidth}{0.4mm}}

\author{Sharadind Peddiraju\IEEEauthorrefmark{1}, Srini Rajagopal
\thanks{Sharadind Peddiraju and Srini Rajagopal are with Amazon Web Services Inc.}
}

\begin{document}
\maketitle
\begin{abstract}
  The challenge of creating domain-centric embeddings arises from the abundance of unstructured data and the scarcity of domain-specific structured data. Conventional embedding techniques often rely on either modality, limiting their applicability and efficacy. This paper introduces a novel modeling approach that leverages structured data to filter noise from unstructured data, resulting in embeddings with high precision and recall for domain-specific attribute prediction. The proposed model operates within a Hybrid Collaborative Filtering (HCF) framework, where generic entity representations are fine-tuned through relevant item prediction tasks. Our experiments, focusing on the cloud computing domain, demonstrate that HCF-based embeddings outperform AutoEncoder-based embeddings (using purely unstructured data), achieving a 28\% lift in precision and an 11\% lift in recall for domain-specific attribute prediction.
\end{abstract}

\noindent\rule{\linewidth}{0.4mm} 
\noindent\rule{\linewidth}{0.4mm}

\section{Introduction}
The effectiveness of machine learning models is largely determined by the accuracy of the data used during training. In various applications, such as customer targeting or behavior prediction, there is often an abundance of unstructured data on companies that might not be domain specific. Whereas, domain specific structured information is relatively less availabled and often highly sparse to extract meaningful signals. . This raises an important question: how can a B2B business tailor ML models to learn from the abundant unstructured data, and customize it to their specific business domain? 
Similar problems exists in multiple other fields of business. For example, for movie ticketing platform, a viewer may have multiple domain related information including their social media posts that is abundant, with business specific data on purchases they have made. Most embedding creation methods today, rely primarily on one of the two types of content described above.

In this paper, we discuss a Hybrid Deep Collaborative filtering model 
that aims to distill knowledge from commonly available unstructured data sources along with business specific structured data to create item embeddings that represent the item domain-level interactions. In the rest of the paper, we use the interaction of companies and technologies as examples, and describe potential data that helps to solve the problem of technology recommendation for companies, but the hybrid deep collaborative filtering model we describe is agnostic to the domain it is used in. In this example, by trying to understand who a customer is, and what their technological needs are, several above-mentioned disadvantages can be well addressed. The work in this paper will solve that problem by condensing data on company technology product consumption into embeddings that can easily be consumed in downstream tasks. In addition, the method discussed in the paper enables adding additional layers of information including 10k filings, wikipedia, and blog postings, making these embeddings richer with other information that is not captured by other third-party data. This approach successfully creates a knowledge representation of a company’s tech stack needs and outperforms other generic models.

The rest of the paper is organized as follows. First, we introduce the problem space and preexisting
science work in the field of work. In section 3, the proposed modeling architecture is introduced with
a clear breakdown of individual modeling components. Sections 4 and 5 of the paper, discuss the data
used in training a model, the experimentation setup, and model training performed over the proposed
architecture. 
\section{Related Work}
\subsection{Creating Domain Representations}
 Domain representations have been primarily developed to study consumption traits of companies or legal entities. Prior work in company representations has focused heavily on the industry classification of companies. Even though there are standard industry classification Standard Industrial Classification (SIC), North American Industry Classification System (NAICS), and the Thomson Reuters Business Classification (TRBC), these are primarily human-controlled classification systems. There has been work on supervised learning methods using unstructured data.\citep{8257920} develop a deep learning model to classify industries based on text from EverString’s database.  
\citep{Pierre2001OnTA} performs industry classification of websites based on their textual content. In more recent work, \citep{WU2023109849} classifies industries based on companies’ supply chain networks using graph neural network methods

\citep{Husmann2020CompanyCU} used unsupervised learning methods to classify company industries. They argue that low-dimensional data resulting from the application of t-SNE can be either provided to experts for visual and exploratory data analysis apart from just creating clusters. \citep{tagarev-etal-2019-comparison} compares several unsupervised and semi-supervised methods over DBpedia company descriptions. to classify industries. However, none of these works create a representation of companies that is closely related to their potential usage of cloud technologies. The embeddings created by \citep{Husmann2020CompanyCU} are useful in categorizing companies, but they rely more on the generic description of companies which is not fine-tuned to work on cloud technologies. All the methods that create embeddings above are unsupervised and have limited utility in specific environments like Cloud computing use cases.

\subsection{Embedding Based Learning}
Embedding-based learning has become a popular technique for extracting learned representations of users in machine learning frameworks. This approach involves creating a low-dimensional vector representation of users, where each dimension represents a single or group of feature(s) and trait(s). Techniques popularly used in the embedding generation, such as Deep Collaborative Filtering (\textbf{DCF}), Graph-Neural Networks (\textbf{GNN}) \citep{scarselli2008graph}, Generative Models (\textbf{GANs}) \citep{grover2016node2vec}, and Auto-encoders (\textbf{AE}) \citep{sutskever2014sequence}, come with trade-offs that directly impact the amount and quality of information retained.

\textbf{AE} models are notably effective for processing textual data  \citep{hinton2006reducing}, whereas \textbf{DCF} and \textbf{GNN} models excel in handling token-based data \citep{kipf2016semi}. However, these approaches are often computationally intensive and inadequate in addressing the cold-start issue. Specifically \textbf{GNN} models require a comprehensive prior of user interactions within a network to mitigate the cold-start problem. In contrast,  \textbf{DCF} models can operate with a less stringent prior, leveraging numerical and demographic side features to enhance the user profile. This flexibility and efficiency of  \textbf{DCF} models in handling user data with minimal prior information guided our focus toward refining the \textbf{DCF} methodology over other embedding generation techniques.

\section{Modeling Architecture}

In this paper, we propose a novel hybrid modeling architecture that takes advantage of both auto-encoders (\textbf{AE}) and Deep collaborative filtering (\textbf{DCF}). As discussed in section 2.2, Auto-encoders primarily use unstructured data to create embeddings whereas \textbf{DCF} is used with structured data that is generally represented as a matrix. The modality of data that is usable in each of the frameworks could be seen as a shortcoming that could be complemented by each other. This is where, \textbf{HCF} architecture adds its novelty. \textbf{HCF} is designed to consume both unstructured free-form text and structured data about a company to create an embedded representation of provided data. As \textbf{HCF} creates these vector embedding representations of a company, the architecture distills information from the unstructured data and retains information that is relevant to the structured data provided.
\subsection{Hybrid Collaborative Filtering (HCF)}

\textbf{HCF} draws conceptual framework from the work of \citep{erhan2010does} on unsupervised pre-training followed by supervised fine-tuning. In a nutshell, unsupervised pre-training layers introduce a useful prior for deep learning models and help the supervised task learn layer weights efficiently leading to a faster loss convergence. The effect unsupervised layers have is close to that of a regularization layer. It does so by restricting the parameter space to particular regions, those that correspond to capturing the structure of the input data.

\textbf{HCF} decomposes training into two stages to maintain the frameworks of  \citep{erhan2010does}, where the first stage is an Auto-encoder framework that creates a vectorized representation of unstructured free-form text. In the second stage, domain-specific information is used to fine-tune the company representations from stage 1. Fine-tuning is optimized by limiting loss function decay on attributes that clearly outline a particular domain helping distill and retain only relevant information. 

\subsection{Stage 1: Unsupervised Pre-Train }

Textual data can be a rich source of prior for companies given the intrinsic relationships between token groups and their contextual usage can be captured. Most of the publicly available data on companies is available as free-form text either through blogs, articles, or wikipedia pages. Large language models that are trained solely on free-form text serve as a testament to the amount of information available publicly on all kinds of entities. The goal of Stage 1 is purely to vectorize available text data to the best of our ability. While rudimentary techniques such as TF-IDF  \citep{salton1988term} and N-grams \citep{mikolov2013efficient} try to capture the contextual presence of tokens to a fair degree, are limited by the vocabulary at training and are brittle in terms of information captured. Using Large language models like Open AI’s ChatGPT or Anthropic’s Claude models to create text vectors could be overkill in our use-case. These models, due to the sheer size of data at training, might create highly generalized vectorization that stands unaffected with fine-tuning at a lower scale. 

Hence, we choose \textbf{\emph{BERT}} (Bidirectional Encoder Representations from Transformers)] \citep{devlin2018bert} to create company vectors as it provides a good balance in performing as a transformer that creates an unsupervised company embedding and associating provided text with a holistic vocabulary. \textbf{\emph{BERT}} has been proven quiet powerful in retaining provided information, thanks to its bi-directional parsing of tokens. To integrate \textbf{\emph{BERT}} into our model, we use it to transform an input matrix of size \textbf{\emph{MxN}}, where \textbf{\emph{M}} represents the number of companies and \textbf{\emph{N}} represents the number of tokens in textual description of the company. \textbf{\emph{BERT}} generates an output $\overrightarrow{E}$, as an embedding of dimension 768 (default dimensions of \textbf{\emph{BERT}}) 

\[\overrightarrow{E} = \textbf{BERT}(M x N)\]

Generated embeddings can then be consumed by other models for training while ensuring that the contextual information is preserved and utilized effectively \citep{reimers2019sentence}. In the development of \textbf{HCF}, we use \textbf{\emph{BERT-base-uncased}} \footnote{https://huggingface.co/google-bert/bert-base-uncased} which is a 110M parameter model and is trained in a self-supervised fashion over Wikipedia and book corpus data. This version of BERT focusing on Masked language modeling (MLM) is particularly suitable for use-cases such as the one at hand, where company information can be partially available.

\subsection{Stage 2: Supervised Fine-Tuning}
In the second stage, HCF implements a Deep Collaborative Filtering algorithm. Collaborative filtering is an embedding-based model that uses matrix factorization to learn numerical representations of items, words, tokens, or entities and users, companies, or individuals. When a collection of $M$ companies is represented as an embedding $E_c \in R^{m\times d}$ and embedding $E_i \in R^{n\times d}$ represents $N$ products across $d$ dimensions, the dot product of $E_c$ and $Ei$ creates an interaction matrix of size $MxN$ capturing the relative interactions of all $M$ companies with $N$ items In equation 1 this interaction matrix is denoted by $\hat{H}^{(m\times n)}$. In \textbf{DCF}, this dot product is used as the initialization step of the neural network, followed by multiple fully connected layers, drop out layers and regularization layers. In each iteration, the factorization machine updates the expectation of the underlying $E_c$ and $E_i$ embeddings to minimize the objective function \citep{rendle2010factorization}. We represent each hidden layer as $L_h$ with an activation function $f$ provided with a weight matrix $W \in \mathbb{R}^{k \times k}$ and bias $b \in \mathbb{R}^k$. We use ridge (L2) regularization for weight decay. This helps in model generalization on unseen data and penalizes large weights in model parameters. Where $\lambda$ is a hyperparameter controlling the strength of regularization. 
\begin{equation}
    E_c^{(m\times d)} \cdot [E_i^{(n\times d)}]^T \approx \hat{H}^{(m\times n)}, 
\end{equation}
\begin{equation}
\begin{aligned}
L_h = f(Wz + b) ,
\end{aligned}
\end{equation}
\begin{equation}
\begin{aligned}
\text{Reg}(E_c, E_i) = \lambda\left(\sum_u |E_{c,u}|^2 + \sum_i |E_{i}|^2\right), 
\end{aligned}
\end{equation}

We use Huber loss function as this penalizes smaller residuals quadratically and linearly on larger residuals. Making the model less sensitive to outliers, this loss function is suited for the modeling use-case at hand where outlier interactions are expected. Huber Loss is defined as below,
\begin{equation}
L{\delta}(y, f(x)) = 
    \left\{
    \begin{array}{ll}
      0.5(y-f(x))^2, & |y-f(x)|\leq \delta.\\
      \delta \cdot (|y-f(x)|-0.5\delta), & \mbox{otherwise}.
    \end{array}
  \right.
\end{equation}

\section{Experimental Setup \& Model Training}
\subsection{Data sets}
To validate the model architecture we conduct experiments with entities and items relevant to cloud services domain. Where companies and business are picked as unique entities and the tech-stack needed by companies to host their business operation are represented as items. We use a combination of 4 data sets, both structured and unstructured, used at different stages of the architecture. Stage 1 uses publicly available information on companies found through Wikipedia, SEC 10-k filings and company descriptions generated through third party data sources. In Stage 2, where we distill information to retain domain specific information, we use a binary table of technology products used by a company. This data source is a combination of information sourced from company websites, job postings, strategic reports and surveys. Through these techniques we are able to create a data source of 10,000 tech products used by companies.

\subsection{Comparative Methods and Experiments Design}

Quantitative performance of the embeddings created through \textbf{HCF} architecture is pegged to the task of tech-product likelihood estimation for a given company. As this is also the objective goal of the fine-tuning task in stage-2 of the model, accuracy of model performance can be directly attributed to the information retained by the embeddings about a customer. Performance evaluation is baselined against each of the independent stages of the proposed architecture. Stage 1, where we use publicly available information on companies with no further fine-tuning, is used to create a embeddings and is tasked to predict likelihood of tech products through a Bayesian Probability Density Based Model (\textbf{BPDM}) \citep{rendle2012bpr}. For a second baseline, we create an independent stage 2 model where only the tabular data is used to train a model to predict tech-product likelihood. In order to execute an objective evaluation, we compare proposed model performance against a (\textbf{BPDM}) and a Memory-Based Collaborative Filtering (\textbf{Mem.CF}) \citep{sarwar2001item} models trained purely on tabular data. By comparing \textbf{HCF} model against industry standard models that operate with tabular data and sub-stages of the proposed model we are able to determine the maximum lift in performance brought by a multi-modal architecture. We evaluate model performance and subsequently the performance of embeddings through Area Under Curve (\textbf{AUC}) of Precision-Recall curves. The highly unbalanced character stemming from the sparity of the tabular data used makes precision-recall curves more apt in performance measurement as these metrics focus only on the positive call (minority class in our data) and do not penalize over true negative. This make Precision-Recall curve a better representation of model performance than standard ROC curves. 

While quantitative evaluation attests to the performance of the model, a more general understanding of how the embedding cluster and represent customers is important to capture the usability of the embeddings in down-stream tasks. Hence, in section 5.2 we examine a network graph of companies and their neighboring communities. We look at top-10 similar companies for a given  company and also examine the top-10 technologies identified to be popular within the sampled communities from a larger company network graph. This evaluation is in-line with qualitative evaluations performed on recommendation systems and embedding based modeling.

\subsection{Training \& Hyper-parameter Tuning}
For each experiment, we divided the data into train, validation and test sets in the ratio of 70\%, 15\% and 15\% respectively, following a standard model training and testing setup. We use two embedding layers for feature extraction in pre-training, and 5 hidden layers coupled with 3 drop-out layers and soft-max regularization to train the model. 
We used 8 NVIDIA A10G Tensor Core GPUs for fine-tuning task. We selected the best group of parameters based on the performance on validation data set, then retrain a model with the identified parameters and report results on test set. We set the hidden units of each hidden layer as 512, 256, 128,  70 and 30. The dropout ratio for each dropout layer is set to 0.3, 0.3, 0.2 and 0.2. We select Adam as our optimizer, and the learning rate is set to 0.018 with a batch size of 512*100. These parameters are a result of hyper-parameter tuning jobs performed through SageMaker HPO jobs \footnote{https://docs.aws.amazon.com/sagemaker/latest/dg/multiple-algorithm-hpo.html} using a Bayesian search algorithm \citep{yu2020hyper} on all model hyper-parameters in an effort to minimize over the loss function (Huber Loss). The final model as shown below in Figure 1 where tech\_emb is the randomly initiated tech-product embeddings, comp\_emb is the company embeddings generated from stage-1 of the framework and dot\_product operation creates the effective company-product matrix. Our resultant model has close to ~100M trainable parameters and takes over 8 hours to train over 100k companies. 

\begin{figure}[h]
    \centering
    \includegraphics[scale=0.25]{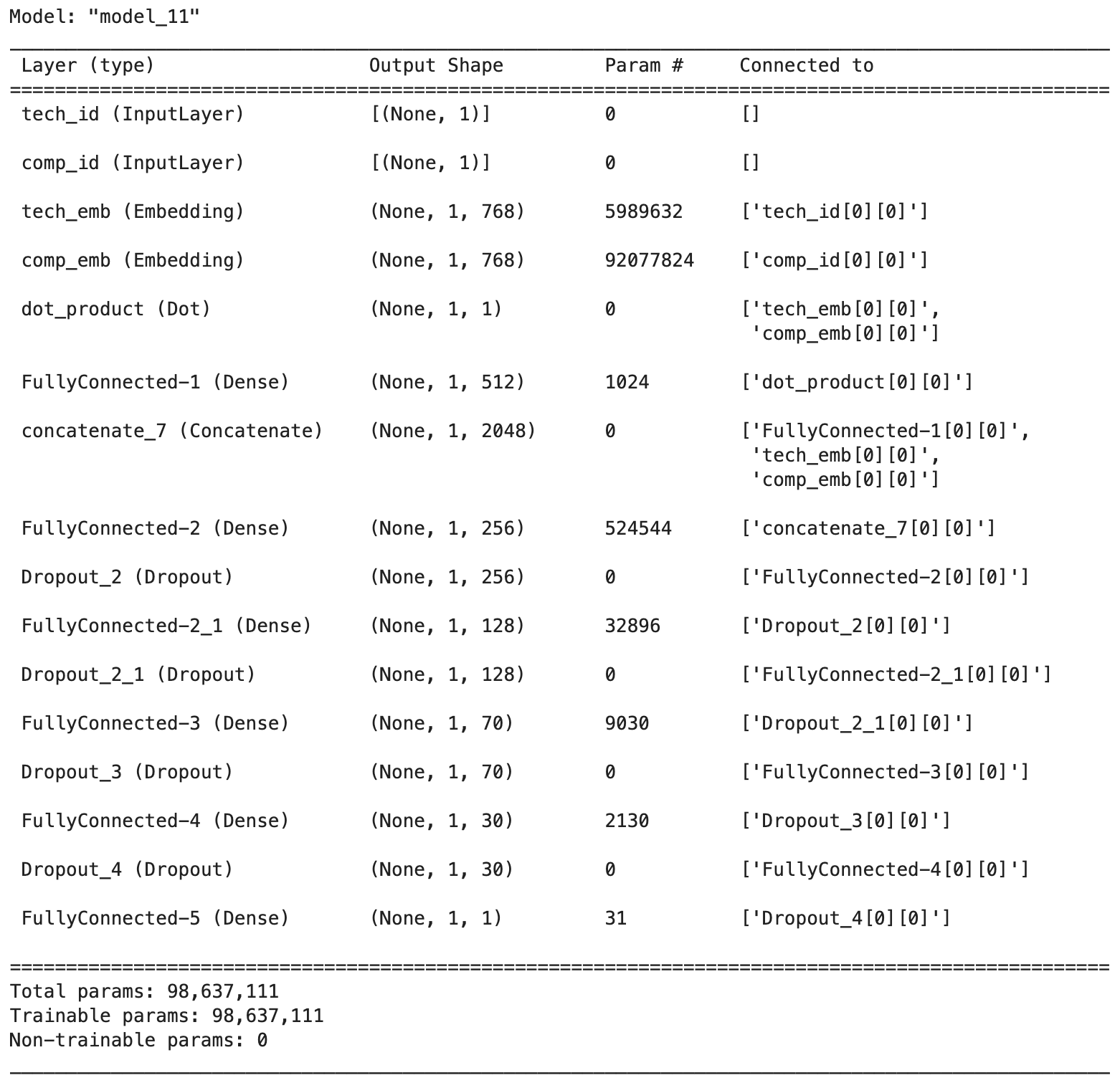} 
    \caption{HCF Model Summary}
\end{figure}

\section{Results \& Discussion} 
\subsection{Quantitative Evaluation: Tech Product Likelihood Prediction}

In our experiments, results outlined in \emph{Table 1}, \textbf{HCF} framework outperforms both the baselines approaches. \textbf{HCF} provides a 36\% lift in precision with \textbf{BPDM} modeling approach and 14\% lift compared to a standard collaborative filtering approach. The lift in performance is consistent across recall and AUC of precision-recall curves. 

\begin{table}
\caption{Tech Product Likelihood Prediction}
\begin{tabular}{cccc}
\toprule
\thead{Model Type} & \thead{Precision} & \thead{Recall} & \thead{PR-RE AUC} \\
\midrule
BPDM & 42\% & 60\% & 45\% \\
Mem.CF & 64\% & 67\% & 65\% \\
Independent Stage 1 & 50\% & 67\% & 57\% \\
Independent Stage 2 & 69\% & 55\% & 64\% \\
\midrule
\textbf{HCF} & \textbf{78\%} & \textbf{78\%} & \textbf{82\%} \\
\bottomrule
\end{tabular}
\end{table}
In the performance evaluation of independent stages of \textbf{HCF}, \textbf{Independent Stage 1} asserts to the Auto-encoder model that takes into account only textual data to predict customer likelihood to tech-products and \textbf{Independent Stage 2}, Deep Collaborative Filtering, module uses only the structured data available at training and randomly initializes the company and product embeddings. This is similar to using a DCF model that has no prior on companies and makes inferences purely based on the product co-occurances observed at training. In line 5 of \textbf{\emph{Table 1}}, \textbf{HCF} model uses company embeddings created in stage 1 as an input to stage 2 and fine-tunes stage 1 embeddings as a result of stage 2 training. While each stage of the \textbf{HCF} framework outperforms all the baselines mentioned in \textbf{\emph{Table 1}}, we can see a significant boost in performance when stage 1 is paired with stage 2 through \textbf{HCF}.

%
%



\subsection{Qualitative Evaluation: Company Community Detection}

Embeddings created through \textbf{HCF} architecture are expected to retain the technology domain knowledge of companies. In effect, this representation should function as a way to cluster companies based on their tech needs and business operations. Such clusters would convey a deeper relationship across different companies compared to a generic industry classification field that captures a partial view of a company. To classify companies into communities, we created a graph where companies are the nodes, and the cosine similarity on company embeddings was used to determine the strength of an edge. Using the Greivan-Neuman community detection algorithm edges are sequentially removed based on their relevance to the neighboring nodes. A detailed structure of the algorithm used in community detection of companies using \textbf{HCF} embeddings is presented in Appendix A of the paper. 
\begin{figure}[h]
  \centering
  \includegraphics[scale=0.2,width=\linewidth]{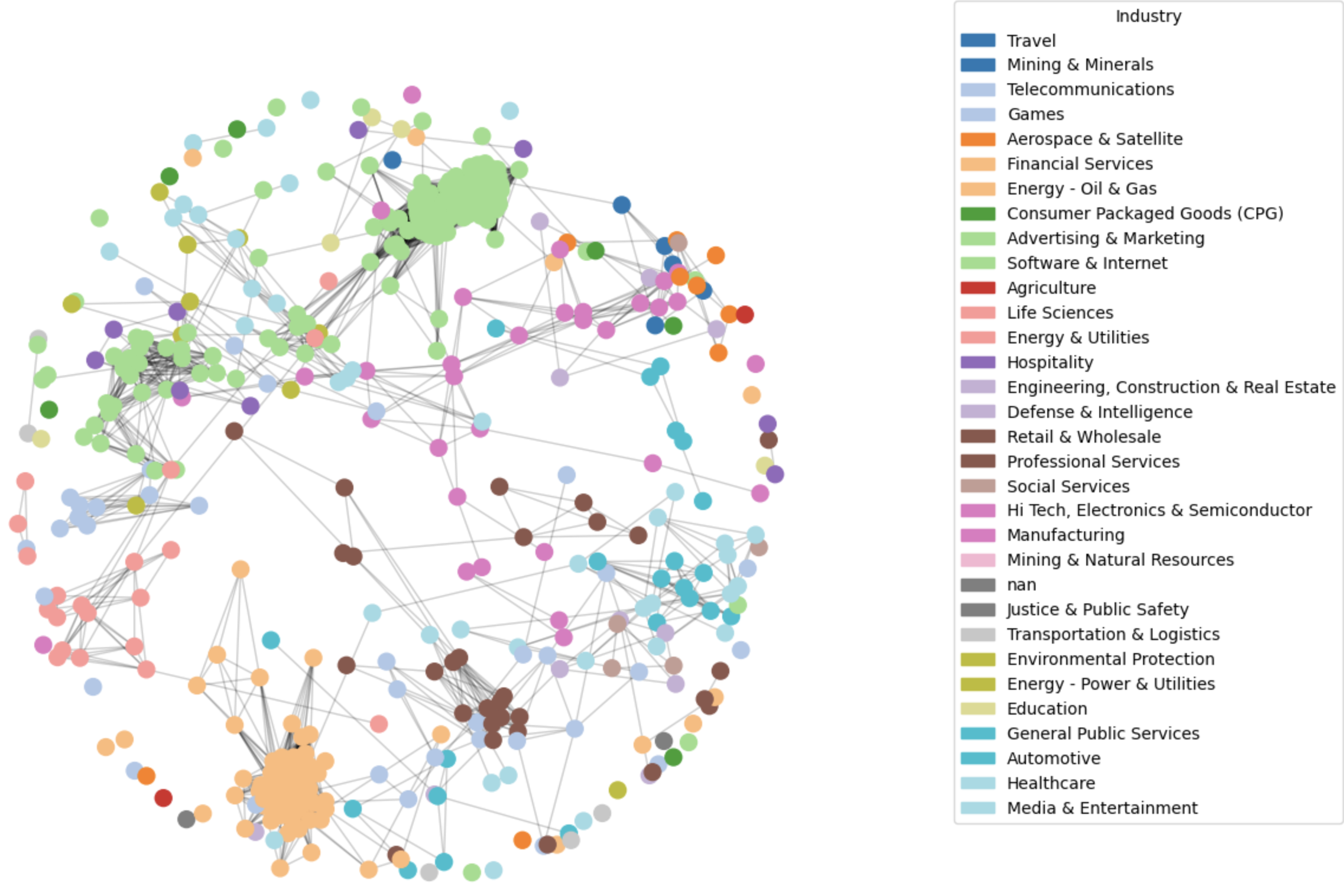}
  \caption{Industry Communities}
\end{figure}

\begin{figure}[h]
  \centering
  \includegraphics[scale=0.25,width=\linewidth]{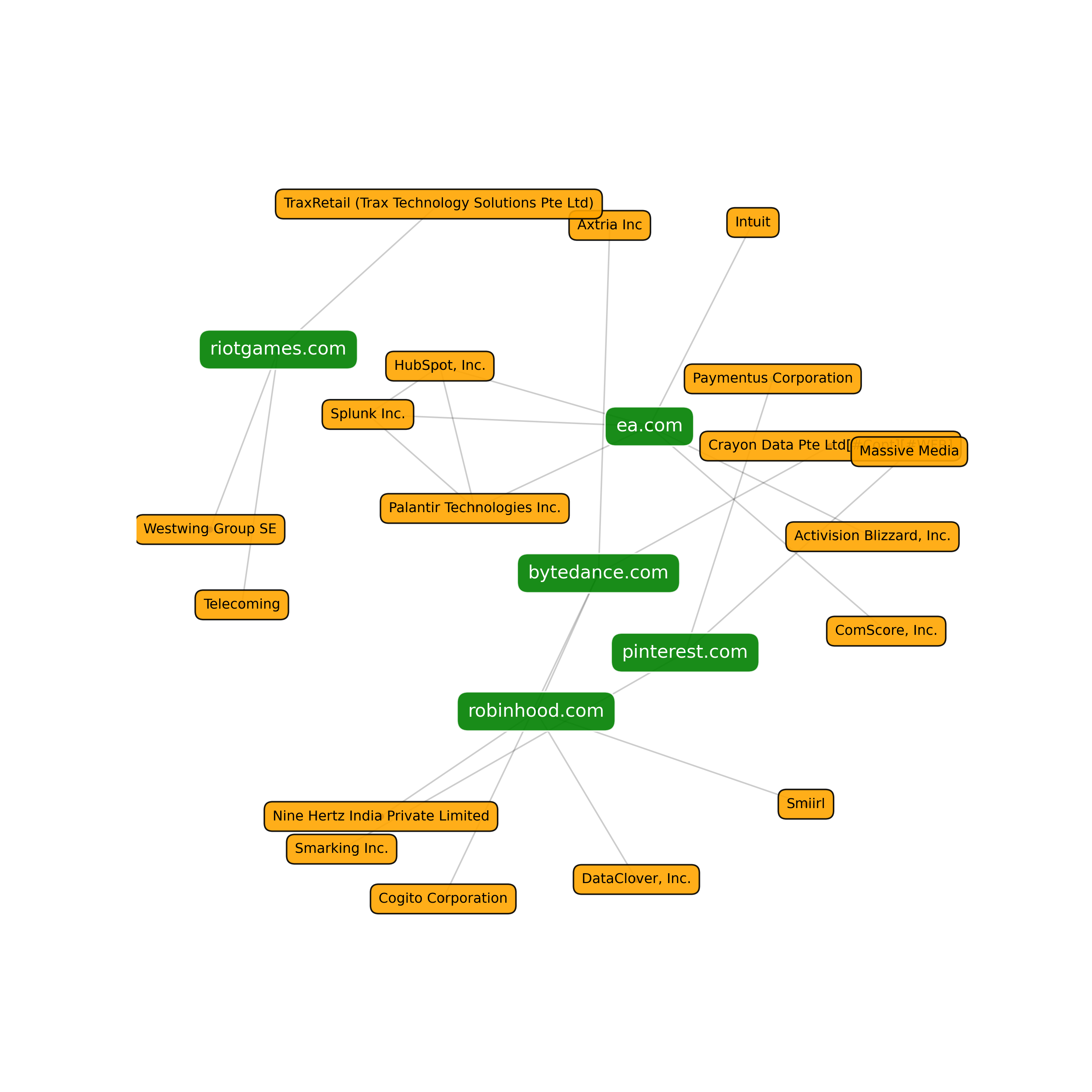}
  \caption{Software Industry Sub-Graph}
  
\end{figure}

Figure 2 shows a random sample of companies classified into communities based on their embedding similarity. Companies from software, advertising and financial services form the most dense communities with most of the constituents of their communities belonging to the same industry. Companies from these industries share technological consumption patterns to their peers in the industry. On the other hand, industries like aerospace, electronics and financial services cluster together into one community, denoting a similar cloud tech-stack consumption though belonging to different industries. 

Figure 3 shows a sub-graph consisting of software industry companies. Interestingly, though EA games, Activision Blizzard, and Riot Games belong to the same community, EA games and Activision Blizzard (both gaming companies) share an edge, but neither is connected to Riot-games. EA and Activision invest heavily in console and PC games developments whereas riot-games customer-facing tech-stack would be more focused on hosting online game play. This can be explained from rudimentary research of how these companies offer products to end customers but upon a close inspection, the Top 5 technologies predicted for neighboring companies of EA, Activision, and Palantir are Teradata, Symantec C++, MySQL, Noe.js, and Docker. Whereas companies connected to the Riot- games have Jenkins, Datawatch, Atlassian-JIRA, MongoDB, and RequireJS as the top 5 technologies used.

\section{Conclusion}
In this paper, we propose \textbf{HCF} - a neural embedding framework for abstracting information from disparate data sources to fine-tune for domain-specific information. We present the effectiveness of the embeddings generated through the framework in the task of predicting user-item interactions and clustering capacity of the embedding to capture cross-industry relationships based on a company's technological needs. The embeddings by themselves also provide a rich abstraction of customer affinity to a variety of technologies, making them effective when consumed directly by auxiliary models.

\bibliographystyle{IEEEtran}
\bibliography{References}

\appendices

\section{Company Community detection throuth HCF embeddings }

\begin{algorithm}[H]
\caption{Company Community Detection via C2V Embeddings}
\begin{algorithmic}[1]
\State \textbf{Input:} Set of company vectors $\mathbf{C}$ from HCF architecture
\State \textbf{Output:} Graph $\mathcal{G}=(V,E)$ representing company communities

\State \textbf{Initialization:}
\State Let $V=\{c_1, c_2, \ldots, c_n\}$ be the set of nodes representing companies
\State Let $E=\emptyset$ be the set of edges representing company relationships
\State Let $\mathbf{sim}: V \times V \to [0,1]$ be the cosine similarity function

\State \textbf{Procedure:}
\State \textbf{for} each pair of companies $(c_i, c_j)$ \textbf{in} $V$ \textbf{do}
\State \ \ \ \ Compute similarity $\sigma_{ij} = \mathbf{sim}(c_i, c_j)$
\State \ \ \ \ \textbf{if} $\sigma_{ij} > \text{threshold}$ \textbf{then}
\State \ \ \ \ \ \ \ \ Add edge $e_{ij}=(c_i, c_j)$ to $E$ with weight $\sigma_{ij}$
\State \textbf{end for}

\State \textbf{Community Detection using Girvan-Newman Algorithm:}
\While{$E \neq \emptyset$}
    \State Calculate edge betweenness centrality $\mathcal{B}(e)$ for all $e \in E$
    \State Find edge $e_{max}$ with the maximum betweenness: $e_{max} = \arg\max_{e \in E} \mathcal{B}(e)$
    \State Remove $e_{max}$ from $E$: $E = E \setminus \{e_{max}\}$
    \State Recalculate connected components $\mathcal{C}$ of $\mathcal{G}$
\EndWhile

\State \textbf{Visualization:}
\State \textbf{Let} $\mathcal{G}_{sub}$ be the subgraph of $\mathcal{G}$ containing a sampled subset of companies
\State \textbf{Let} $N_c$ be the set of top closest neighbors for each sampled company in $\mathcal{G}_{sub}$
\State Display $\mathcal{G}_{sub}$ and $N_c$ in Figure 2, illustrating the community structure

\end{algorithmic}
\end{algorithm}

\section{Data Ablation Study \& Performance Charts}
To understand the lift in performance from different data sources used in the modeling we perform an ablation study. Currently, we use company descriptions form Wikipedia and other public sources as the source of company context (\textbf{CC}). We use description of products and technologies to further enhance the company context. We extend the ablation study by modulating the number of technologies used at training to understand the impact of the filtering techniques (data sparsity vs. information richness trade off discussed in Section 4.1) used in the development of the main model.  Table 3 provides the performance of each of the modeling components following the notation : \textbf{\emph{CC}} for company context,  \textbf{\emph{Tech Desc}} for descriptions of technologies used, \textbf{\emph{Tech Prod}} for occurrences of technologies used by a company. 
\begin{table}
\scriptsize
\caption{ \label{Tabel 2} Ablation Study - Accuracy Metrics}
\begin{tabular}{ccccc} 
\toprule
 \thead{\scriptsize Model Type} & \thead{\scriptsize Feature Sets} & \thead{ \scriptsize Technology\\\scriptsize Count} &  \thead{\scriptsize Precision} &  \thead{\scriptsize Recall} \\
\midrule
  & CC & - & 49.5\% & 67\%  \\
 BERT AE & \thead{\scriptsize CC \& Tech Desc} & -  & 49.6\% & 67.6\% \\
\midrule
  &  & 1K & 68\% & 52\%  \\
 DCF & \thead{\scriptsize Tech Prod} & 5K & 67\% & 60\%  \\
  &  & 10K & 69\% & 55\%  \\
\midrule
 HCF & \thead{\scriptsize Pre-train: CC\\ \scriptsize Fine-tune: Tech Prod} & 1K & 65\% & 51\%  \\
  &   & 5K & 68\% & 55\%  \\
\midrule
 HCF &  \thead{\scriptsize Pre-train: CC \& Tech.Desc \\ \scriptsize Fine-tune: Tech Prod} & 1K & 66\% & 55\%  \\
  &  & 5K & 71\% & 58\%  \\
\midrule
  &  & 1K & 72\% & 65\%  \\
 HCF & \thead{\scriptsize Pre-train: CC, Tech.Desc \\ \scriptsize Fine-tune: Tech Prod}  & 5K & 74\% & 68\%   \\
  &   & 10K & 78\% & 78\%  \\
\bottomrule
\\
\end{tabular}
\end{table}

\begin{table}[h]
\scriptsize
\caption{ \label{Tabel 3} Ablation Study - Area Under Curves}
\begin{tabular}{ccccc} 
\toprule
 \thead{\scriptsize Model Type} & \thead{\scriptsize Feature Sets} & \thead{ \scriptsize Technographic \\ \scriptsize Count} &   \thead{\scriptsize PR-RE\\ \scriptsize AUC} &  \thead{\scriptsize ROC\\ \scriptsize AUC} \\
\midrule
  & CC & - &  57\% & 74\% \\
 BERT AE & \thead{\scriptsize CC \& Tech Desc} & -    & 57\% & 75\%\\
\midrule
  &  & 1K & 62\% & 76\% \\
 DCF & \thead{\scriptsize Tech Prod} & 5K &  65\% & 79\% \\
  &  & 10K &  64\% & 78\% \\
\midrule
 HCF & \thead{\scriptsize Pre-train: CC\\ \scriptsize Fine-tune: Tech Prod} & 1K &   60\% & 73\% \\
  &   & 5K &   71\% & 85\% \\
\midrule
 HCF &  \thead{\scriptsize Pre-train: CC \& Tech.Desc \\ \scriptsize Fine-tune: Tech Prod} & 1K &   61\% & 74\% \\
  &  & 5K &   70\% & 83\% \\
\midrule
  &  & 1K &   74\% & 86\% \\
 HCF & \thead{\scriptsize Pre-train: CC, Tech.Desc \\ \scriptsize Fine-tune: Tech Prod}  & 5K &   80\% & 87\% \\
  &   & 10K &  82\% & 82\% \\
\bottomrule
\\
\end{tabular}
\end{table}


\section{Data Sparsity Vs Information Richness Trade-off}

Stage 2, where the model deals with product co-occurrence presents us with the challenge of increasing sparsity as the size of the products covered in the matrix increases. High sparsity in this data can lead to heavy computational needs, weak generalization, and bias in models. On the other hand, having a product that is used by almost every single company in the data set would add very little value in fine-tuning tasks. Hence, In selecting the products used for fine-tuning we create a heuristic metric to filter products used in stage 2. 

\[\text{occurrence density} (\rho) = \frac{\text{number of companies using product}}{\text{total number of companies}}\]
\[\text{product selection flag} = \begin{cases} 
1 & \text{if } 0.1 \leq \rho \leq 0.85 \\
0 & \text{otherwise}
\end{cases}
\]

This filter helps maintain information richness along with managing data sparsity. As this filter excludes highly frequent products ( used by $\geq$ 85\% of companies) or products that are used by a small set of companies ( $<$ 10\% ), helps reduce data sparsity while retaining most of the products. Through this filtering, 5032 technology products from a larger pool of 10000 products were retained for fine-tuning. 

\bibliographystyle{IEEEtran}

\end{document}